\documentclass[10pt,twocolumn,letterpaper]{article}

\usepackage{cvpr}              % To produce the CAMERA-READY version
\definecolor{cvprblue}{rgb}{0.21,0.49,0.74}
\usepackage[pagebackref,breaklinks,colorlinks,allcolors=cvprblue]{hyperref}

\graphicspath{{figs/}}
\usepackage{xspace}
\usepackage[table,xcdraw,dvipsnames]{xcolor}
\usepackage{color}
\usepackage{array}
\usepackage{comment}

\usepackage{booktabs}  
\usepackage{multirow}   
\usepackage{graphicx}   
\usepackage{amsmath}  

\usepackage{xcolor}
\usepackage{pifont}

\usepackage[normalem]{ulem}
\useunder{\uline}{\ul}{}

\def\paperID{39826} % *** Enter the Paper ID here
\def\confName{CVPR}
\def\confYear{2026}

\newcommand{\ours}{FlexMem\xspace}

\title{Scaling the Long Video Understanding of Multimodal Large Language Models via Visual Memory Mechanism}

\author{
    Tao Chen\textsuperscript{1}\:  
    Kun Zhang\textsuperscript{1}\: 
    Qiong Wu\textsuperscript{1}\: 
    Xiao Chen\textsuperscript{1}\: 
    Chao Chang\textsuperscript{2}\: \\
    Xiaoshuai Sun\textsuperscript{1}\:\: 
    Yiyi Zhou\textsuperscript{1}\setcounter{footnote}{1}\thanks{Corresponding author: zhouyiyi@xmu.edu.cn.}\:\:\: 
    Rongrong Ji\textsuperscript{1} \\ 
    \textsuperscript{1}Key Laboratory of Multimedia Trusted Perception and Efficient Computing,\\
    Ministry of Education of China, Xiamen University, 361005, P.R. China. \\
    \textsuperscript{2}National University of Defense Technology, 230000, P.R. China. \quad \\ 
}

\begin{document}
\maketitle
\begin{abstract}

Long video understanding is a key challenge that plagues the advancement of \emph{Multimodal Large language Models} (MLLMs). In this paper, we study this problem from the perspective of visual memory mechanism, and proposed a novel and training-free approach, termed \emph{Flexible Memory} (\textbf{FlexMem}).  In principle, FlexMem aims to mimic human behavior of video watching, \emph{i.e.}, continually watching video content and recalling the most relevant memory fragments to answer the question. In this way, FlexMem can help MLLMs achieve video understanding of infinite lengths, unlike previous methods that process all video information at once and have input upper-limit. Concretely, FlexMem first consider the visual KV caches as the memory sources, and realize the effective memory transfer and writing via a dual-pathway compression design.  Afterwards, FlexMem also explores different memory reading strategies for the diverse video understanding tasks, including the popular streaming one.
To validate FlexMem, we apply it to two popular video-MLLMs, and conduct extensive experiments on five long video and one streaming video task. The experimental results show that on \textbf{a single 3090 GPU}, our FlexMem can achieve obvious improvements than existing efficient video understanding methods and process more than \textbf{1k frames}, which also helps the base MLLMs achieve comparable or even better performance than SOTA MLLMs on some benchmarks, \emph{e.g.} , GPT-4o and Gemini-1.5 Pro. Our code is released at: \href{https://github.com/city1517/FlexMem}{\textcolor[rgb]{0,0.2,0.5}{FlexMem}}.
\end{abstract}

\section{Introduction}
\label{sec:intro}

Recent years have witnessed the remarkable progress made by \emph{Multimodal Large Language Models} (MLLMs)~\cite{Zhu0SLE24, zhou2019plenty,zhou2021trar,TongLZLS0SJ25,LuoZ0ZSJ25,abs-2411-19628} towards effective vision-language understanding. Despite the great success, long video understanding is still a main obstacle for existing MLLMs mainly due to the difficulty of processing excessive long video frames~\cite{RenYL0H24,0004LJJCSSL24}. In addition to high computation complexity, the large number of visual tokens from long videos can easily exceed the upper limit of the sequence length of existing MLLMs~\cite{abs-2504-06835,GanWSWGN23,abs-2410-17434}, \emph{e.g.},  more than 200$k$ for 1024 video frames~\cite{0080ZGZ00ZZL0L25}, resulting in both performance degradation and expensive memory overhead~\cite{0080ZGZ00ZZL0L25,ChenXLHZLFTYLHY25}.

\begin{figure}[t]
  \centering
  \vspace{-10pt}
   \includegraphics[width=\columnwidth]{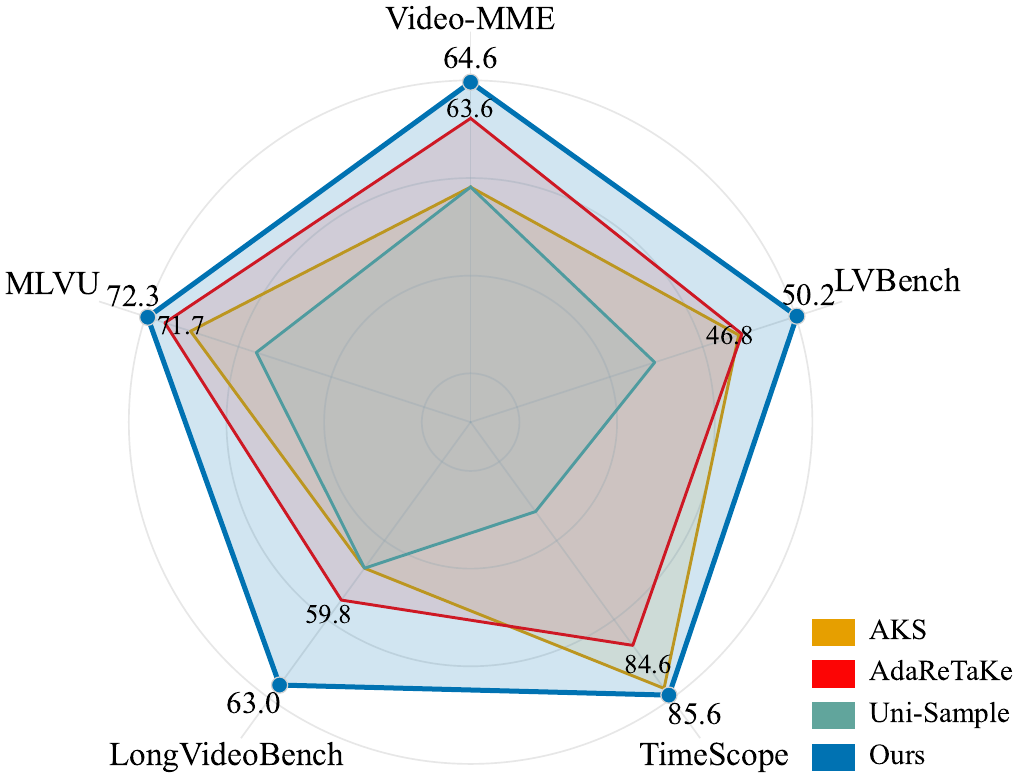}
   \caption{Comparison between FlexMem (ours) and existing efficient video understanding methods for MLLMs on five benchmarks. All methods are run on the same device of one 3090 GPU, and our FlexMem presents obvious performance gains.}
   \label{fig:preview}
    \vspace{-15pt}
\end{figure}

To tackle this issue, recent efforts~\cite{0004LJJCSSL24,TangQXTJY25,FanMWDLGL24,WengHHCZ24} are devoted to efficient long video understanding for MLLMs. One popular solution is to adopt \emph{retrieval augmentation generation} (RAG) based strategies to select key video information for MLLMs~\cite{abs-2411-13093, TangQXTJY25}, drawing on the successful experience of LLMs~\cite{ShiMYS0LZY24,AsaiWWSH24}. Concretely, RAG methods regard the whole video as a knowledge base, and then find out the question-related key frames (or clips) as the input of MLLMs, thereby avoiding the processing of all frames. Although effective in video tasks like \emph{needle-in-a-haystack}~\cite{ZhaoLH0YGWCL25}, which requires evident localization from thousands of video frames, RAG methods are still inferior in mastering continual and overall understanding of videos~\cite{abs-2502-01549,LiuJDLWGL25}. In this case, they are still sensitive to memory overhead for more keyframe inputs~\cite{abs-2408-03340,Cao2025}. The other viable solution is to use visual feature compression for the longer input of video frames~\cite{WangSZWCN25, DiYZLZCLHSJ25, YuanYB25}. For instance, Wang~\emph{et al.}~\cite{WangSZWCN25} apply visual \emph{Key-Value} (KV) caches compression to reduce the per-clip footprint, thereby increasing the number of input frames. However, visual compression methods~\cite{WangSZWCN25,abs-2412-20504,ShuLZQ0L0025} still require MLLMs to input all compressed visual features for the final answering, still yielding obvious computation bottlenecks. Overall, existing methods are still hard to strike a trade-off between efficient video understanding and optimal performance.  

In this paper, we study the long video understanding of MLLMs from the perspective of visual memory mechanism~\cite{FanMWDLGL24,0004LJJCSSL24,abs-2406-08085}.
Specifically, we aim to help MLLMs to be able to watch videos continuously, form visual memories and answer questions based on relevant memory fragments, just like a human.
In this way, MLLMs can answer the question without having to using all information, \emph{i.e.}, breaking the input limit of the final prediction, while also being capable of handling different question types, \emph{e.g.}, the global and general ones. 
More ideally, this memory mechanism should be also independent to MLLMs' structure and training, and can be a plug-and-play component that directly applied to MLLMs without great structure tweaks.

However, achieving the above target still encounters several key challenges. The first one is how to effectively encode memory fragments. While some recent works use KV caches as the viable representations~\cite{abs-2506-11418, abs-2502-01068,WangSZWCN25}, we think that the memories for video MLLMs should not be only highly compressed but also transferable and continuous, thereby handling different types of video tasks, as discussed above. Secondly, the effective reading of memory is also critical. One intuitive solution is to leverage the MLLM's cross-modal attention during encoding to judge the relevance of memories. However, in scenarios with multiple questions or streaming QA~\cite{NiuLMGZHDDD0ZZC25, abs-2411-03628}, the repeated encoding of video clips and answers will incur excessive computation overhead. In this case, the design of effective and efficient visual memory mechanism for MLLMs is still a intractable problem.

To address these challenges, we propose a novel and training-free visual memory mechanism for video-MLLMs, termed \emph{Flexible Memory} (FlexMem). Concretely, FlexMem resorts to \emph{Key-Value} caches of visual tokens as the MLLM's memory representations, similar to some existing compression-based works~\cite{abs-2505-15269,abs-2412-20504}. In practice, we also introduce a novel \emph{dual-pathway compression} design that can greatly reduce the memory sizes while ensuring the continuity of each memory snippet. 
In terms of memory reading, FlexMem is also equipped with a novel and fast indexing approach in addition to the aforementioned encoding-based one, called \emph{MemIndex}. Via statistically fitting the encoding-based retrieval, MemIndex adaptively select the representative cache layers and tokens to form a much smaller memory index tensor, supporting the fast and flexible memory retrieval.
With these innovative designs, the proposed FlexMem can scale the input frames of MLLMs, thereby significantly enhancing their long video understanding.

To validate \ours, we apply it to two representative video MLLMs, namely LLaVA-OneVision~\cite{0080ZGZ00ZZL0L25} and LLaVA-Video~\cite{ZhangWLLMLL25}, and conduct extensive experiments on a bunch of highly competitive benchmarks. The experimental results not only show the great improvement to video MLLMs, \emph{e.g.}, +32.2\% on TimeScope for LLaVA-Video, but also validate its merits than existing methods for efficient video understanding. For instance, under the same setting of one 3090 GPU, our FlexMem can outperforms the SOTA methods such as AKS~\cite{TangQXTJY25} and AdaRETAKE~\cite{WangSZWCN25} by 3.9\% and 5.2\% on average for LLaVA-Video, respectively.

Overall, our contributions are two-fold:
\begin{itemize}
    \item We study the long video understanding of MLLMs from the perspective of visual memory mechanism, and propose a novel approached termed FlexMem to scale up the input of video frames.
    \item On a set of benchmarks, our \ours can greatly improve the capabilities of base MLLMs and outperform a set of SOTA methods using only one 3090 GPU.
\end{itemize}

\section{Related Work}

\subsection{Video Multimodal Large Language Models}

The rapid advancement of Large Language Models (LLMs) has catalyzed significant breakthroughs in multimodal understanding~\cite{luo2022towards,luo2024moil,luo2024towards}, leading to the emergence of Video Multimodal Large Language Models (Video-MLLMs)~\cite{AlayracDLMBHLMM22,abs-2305-06355,LinYZCNJ024,0001RKK24}. 
Early pioneering works like Flamingo~\cite{AlayracDLMBHLMM22} and VideoChat~\cite{abs-2305-06355} laid the foundation by extending image-based multimodal models with temporal modeling modules, enabling basic video comprehension capabilities. 
Subsequent works such as Video-LLaVA~\cite{LinYZCNJ024} and Video-ChatGPT~\cite{0001RKK24} improve temporal reasoning through unified visual representations and joint image-video training. 
More recent state-of-the-art models like Qwen3-VL~\cite{Qwen2.5-VL} and InternVL3.5~\cite{abs-2508-18265} have achieved remarkable performance improvements by scaling both model parameters and training data. 
However, despite their impressive capabilities, these methods are fundamentally constrained by computational resources and typically process only a limited number of frames, which significantly restricts their applicability to long video understanding scenarios.

\begin{figure*}[ht]
  \centering
   \includegraphics[width=0.85\linewidth]{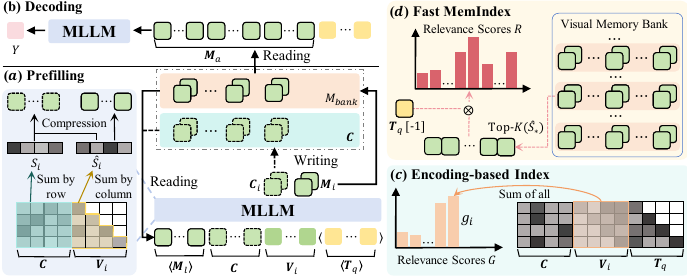}
    \vspace{-5pt}
    \caption{Illustration of the proposed \ours method. (a) \ours is an iterative method, and it encodes two types of compressed memories for each video clip $V_i$, namely \emph{Context Memory} $C_i$ and \emph{Local Memory} $M_i$, based on the metrics of aggregation score $S_i$ and local saliency score $\hat{S}_i$, respectively. 
    $M_i$ is then stored in the \emph{visual memory bank} $M_{bank}$, while the context memory $C$ are used in the iterative encoding step for information propagation. Besides, we can also retrieval some stored $M_l$ as the long-term memory for encoding, while it is optional as well as the text instruction $T_q$.
    (b) The stored memories $M_a$ will be recalled from the memory bank for the decoding of answers $Y$. (c) One intuitive and effective indexing for FlexMem is the \emph{Encoding-based} one, which uses the cross-attention during memory encoding with $T_q$ (a) to reflect the relevance of memories. (d) We also investigate the other fast index method, termed \emph{MemIndex}, based on the compact index tensors for both question and visual memories, of which process is independent to the encoding of memories. Its selection of cache layers and tokens stems from the fitting results of the encoding-based index.}
   \label{fig:overview}
    \vspace{-10pt}
\end{figure*}

\subsection{Long Video Understanding}

To tackle the above challenge, some efforts resort to Retrieval-Augmented Generation (RAG) strategies derived from LLMs~\cite{ShiMYS0LZY24,AsaiWWSH24} to long video understanding. Video-RAG methods~\cite{TangQXTJY25,abs-2502-01549,abs-2411-13093,abs-2508-01546, abs-2510-14032} typically employ a two-stage pipeline, \emph{i.e.}, first retrieving keyframes based on query similarity, then processing them for answer generation.
For instance, AKS~\cite{TangQXTJY25} uses vision-language embedding Models for similarity-based retrieval, while VideoAgent~\cite{FanMWDLGL24} employs an iterative refinement process with LLM-based planning.
However, such retrieval methods face inherent limitations in maintaining temporal coherence and capturing long-range dependencies. These methods often lack important contextual information that spans multiple segments and struggle with queries requiring holistic video understanding. 
Recently, visual compression methods have been extensively studied~\cite{WangSZWCN25,abs-2412-20504,ShuLZQ0L0025}, which maintain compressed features of historical context for comprehensive understanding.
For instance, AdaRETAKE~\cite{WangSZWCN25} designs adaptive allocation modules to determine compression ratios across temporal dimensions and MLLM layers.
Video-XL~\cite{ShuLZQ0L0025} introduces special tokens to summarize the visual information within video fragments. 
Despite these advances, their input context length grows linearly with video duration, limiting their scalability. 
\ours combines the benefits of both paradigms, \emph{i.e.}, maintaining comprehensive visual memories with constant footprint through iterative processing, and reading the most relevant information for answer generation via memory recall mechanism. 

\section{Method}
\label{sec:formatting}

\subsection{Overview}
In this paper, we study the long video understanding of MLLMs from the perspective of visual memory mechanism, and propose a novel and \emph{training-free} approach termed \emph{Flexible Memory} (FlexMem), as depicted in Fig.~\ref{fig:overview}. 

In principle, FlexMem aims to mimic the human behaviors of video watching, \emph{i.e.}, continually browsing video content, forming memories and answering questions based on memory recall. Via this iterative paradigm, FlexMem can help MLLMs break the upper-limit of input length.

In particular, given a long video $V$ and a text instruction $T_q$, existing MLLMs~\cite{abs-2409-12191,abs-2508-18265,ZhangWLLMLL25} normally sample a subset of frames as the visual input, denoted as $V'=\{I_1,\cdots, I_M\}$, due to the limit of input sequence length and memory overhead. The prediction $Y$ is generated according to all input frames and the text instruction:
\begin{equation}
    \centering
    \begin{aligned}
        & \text{MLLM}(I_1,\dots,I_M, T_q) \rightarrow Y.  \\
    \end{aligned}
\end{equation}

In terms of long video understanding, this solution is greatly limited by the number of input frames, leading to suboptimal performance~\cite{WuLCL24,abs-2409-18938}. 
To address this challenge, \ours considers the visual KV caches as the memory sources, and realizes the effective memory transfer and writing via a dual-pathway compression design.

Specifically, we first divide the video into $N$ clips $V=\{V_1,\cdots, V_N\}$. Then, FlexMem lets MLLMs to read video clips iteratively, and its first step is defined by
\begin{equation}
    \centering
    \begin{aligned}
        & \text{MLLM}(V_1, \langle T_q \rangle)\rightarrow M_1, C_1.\\
    \end{aligned}
\end{equation}
where $\langle \cdot \rangle$ is an optional input. $M_1, C_1$ are the compressed local memory and context memory respectively, which are processed by our \emph{Dual-Pathway Compression} (DPC) design. In particular, $M_1$ is written into the visual memory bank $M_{bank}$ for the following memory recall, while $C_1$ is used for the historical video information propagation in the iterative steps. Thus, they are processed by differently. 

After the first step, FlexMem will extend the inputs of MLLMs, which can be defined by 
\begin{equation}
\label{eq:o_iter}
\resizebox{0.9\columnwidth}{!}{$\displaystyle{
\text{MLLM}(\langle M_l \rangle,C_{k-n_s},...,C_{k-1}, V_k, \langle T_q \rangle)\rightarrow M_k, C_k.
}$}
\end{equation}
Here $k$ denotes the current step of memory processing, and $n_s$ is the number of retained context memories. In Eq.~\ref{eq:o_iter}, we give a certain interval of previous context memories to MLLM, thereby achieving the transfer of video information and building the continuity of stored memories. Besides, we also recall some stored $M_l$ from the memory bank as the long-term memory for the better understanding of long historical information.

After watching the whole video, FlexMem will recall the most relevant memory pieces from $M_{bank}$:
\begin{equation}
    \centering
    \begin{aligned}
        & \text{Recall}(M_{bank},T_q) \rightarrow M_i,...,M_{i+n_a-1}. \\
    \end{aligned}
\end{equation}
where $M_i$ is the recalled memory, and $n_a$ is the number of recalled pieces. Lastly, MLLM will use these recalled memories for the final answer prediction:
\begin{equation}
    \centering
    \begin{aligned}
        & \text{MLLM}(M_i,...,M_{i+n_a-1},T_q)\rightarrow Y\\
    \end{aligned}
\end{equation}

In particular, the memory encoding with $T_q$ in Eq.~\ref{eq:o_iter} is optional. According to the \emph{uni-directional attention mechanism} of MLLMs~\cite{VaswaniSPUJGKP17,wu2026not}, the encoding of $T_q$ will not affect the visual memory compression, but can help to record the video-question relevance for the following memory recall. In this paper, we also explore the fast indexing of video memories, \emph{i.e.,} not using $T_q$ during the encoding of visual memories. Besides, FlexMem is an iterative approach, \emph{i.e.}, Eq.~\ref{eq:o_iter}, which can theoretically process infinite-long videos.

\subsection{Dual-Pathway Compression}

To scale long video understanding, FlexMem is equipped with a novel \emph{dual-pathway compression} design for memory compression and transmission. In particular, FlexMem also regards the encoded \emph{Key-Value} caches of visual tokens as the memory source, and effectively compresses them for memory writing and reading. 
Compared with existing KV cache compression methods~\cite{WangSZWCN25,ShuLZQ0L0025,abs-2412-20504}, which progressively encode clips and have input upper-limit, our FlexMem consider visual memory encoding as a iterative process that focuses on information transfer.

Concretely, at the $i$-st step of FlexMem, we will include the recent context memory $C=\{C_{i-k}\}_{k=1}^{n_s}$ into the encoding of current clips, and return the attention matrix $A^l$:
\begin{equation}
\label{eq:kvcache}
\centering
    \begin{aligned}
        & \mathbf{A}^l_v = \text{Attention}([\mathbf{Q}_{V_i}, \langle\mathbf{Q}_{T_q}\rangle], [\mathbf{\hat{K}}_{C},\mathbf{K}_{V_i},\langle\mathbf{K}_{T_q}\rangle]), \\
    \end{aligned}
\end{equation}
where $\mathbf{Q}_{V_i}$ and $\mathbf{K}_{V_i}$ are the query and key vectors of $V_i$ at each layer, and $\mathbf{Q}_{T_q}$ and $\mathbf{K}_{T_q}$ are those of $T_q$.

Recognizing that the role of the visual memory differs between the prefill and decoding stages, we strategically prunes unimportant KVs of $V_i$ based on two attention-based metrics. 
For the prefill stage, the objective is to encode the current clip with a rich understanding of its historical context, \emph{i.e., context memory $C$}.

To approach this target, we measure the importance of a token whether it effectively aggregates information from past context and propagates its own information to subsequent tokens within its clip. 
We define the context aggregation score $s^l_j$ for the $j$-th token in clip $V_i$ as the metric for obtaining its context features $\mathbf{c}_i^l$:
\begin{equation}
\label{eq:s_pre}
\centering
    \begin{aligned}
        & \mathbf{c}_i^l=\{\mathbf{k}_j^l,\mathbf{v}_j^l|s^l_j \in \overset{\alpha_c|V_i|}{\underset{j\in V_i}{\arg\max}} \: s_{j}^l\}, \\
        & \text{where} \:\:s^{l}_j = \sum_{k \in C} a_{jk}^l + \sum_{h \in V_i} a_{hj}^l.
    \end{aligned}
\end{equation}
where $a_{jk}^l$ is attention weight of $A_v^l$ from $j$-th token in current clip to $k$-th token in the historical context.
$k_j^l,v_j^l$ are the key and value vectors of the $j$-th token at the $l$-th layer. 
$\alpha_c$ denotes the compression ratio for context features, and $|V_i|$ is the number of tokens in clip $V_i$.
The context memory $C_i$ of clip $V_i$ is consisted of its KVs from all cache layers, \emph{i.e.}, $C_i=\{\mathbf{c}_i^1,\dots,\mathbf{c}_i^L\}$.

For the decoding stage, the MLLMs aim to answer the text instruction based on the most salient visual evidence. Therefore, the priority at this time is to eliminate redundancy within each clip to retain its most distinctive information. We thus define a local saliency score $\hat{s}^l_j$ to measure the overall influence of a token within its own clip, and use it to obtain compressed visual features $\mathbf{m}_i^l$:
\begin{equation}
\label{eq:s_dec}
\centering
    \begin{aligned}
        & \mathbf{m}_i^l=\{\mathbf{k}_j^l,\mathbf{v}_j^l|\hat{s}^l_j \in \overset{\alpha_s|V_i|}{\underset{j\in V_i}{\arg\max}} \: \hat{s}_{j}^l\}, \\
        & \text{where} \:\: \hat{s}^l_j = \sum_{k \in V_i} a_{kj}^l.
    \end{aligned}
\end{equation}
where $\alpha_s$ is the compression ratio for the stored memory $M_i$, and it includes the compressed caches $\mathbf{m}_i^l$ of all layers of the clip $V_i$, \emph{i.e.}, $M_i=\{\mathbf{m}_i^1,\dots,\mathbf{m}_i^L\}$. 

Overall, FlexMem can iteratively process extra long videos with limited memory overhead, and obtain the stored memory bank $M_{bank}$ for prediction, of which features are rich in visual information and continual in video semantics.

\subsection{Memory Reading}

\subsubsection{Question Encoding based Memory Reading}

In terms of memory reading, one effective solution is to directly uses the cross-modal attention encoded during memory compression, \emph{i.e.}, Eq.~\ref{eq:kvcache}. Based on the superior \emph{vision-langauge} (VL) alignment capability of MLLMs, we can directly use the cross-modal attentions between video clips and question as the metric for memory reading.

Specifically, we compute this relevance score $g_i$ by summing the attention weights from the instruction tokens to the tokens of clip $V_i$ at the prefill stage:
\begin{equation}
\label{eq:s_ret}
\centering
    \begin{aligned}
        & g_i = \sum_{l =3}^L \sum_{j \in T_q}  \sum_{k \in V_i} a_{jk}^l, \\
        &  \text{Recall}(M_{bank}, T_q) = \{M_i|g_i \in \overset{n_a}{\underset{i\in M_{bank}}{\arg\max}} \: g_i\}.
    \end{aligned}
\end{equation}

\begin{table*}[ht!]
\centering
\caption{A comparison of \ours with SOTA methods based on two recent MLLMs across five long VideoQA benchmarks. \emph{Sampled Frames} denote the number of frames sampled from the video used for compression or selection, and \emph{Input Tokens} denote the number of tokens used for question answering. The best and second-best results are shown in \textbf{bold} and \underline{underlined} respectively. $^*$Tested on one A800.}
\vspace{-5pt}
\label{tb:overall_method}
\resizebox{\linewidth}{!}{
\renewcommand{\arraystretch}{0.8}
\begin{tabular}{c|c|c|c|c|c|cccc|cccc}
\toprule
\multirow{2}{*}{\textbf{Method}}  & \multirow{2}{*}{\textbf{\begin{tabular}[c]{@{}c@{}}Sampled\\ Frames\end{tabular}}} & \multirow{2}{*}{\textbf{\begin{tabular}[c]{@{}c@{}}Input\\ Tokens\end{tabular}}} & \textbf{TimeScope} & \textbf{LVBench} & \textbf{MLVU} & \multicolumn{4}{c|}{\textbf{Video-MME}} & \multicolumn{4}{c}{\textbf{LongVideoBench}} \\ \cmidrule(l){4-14} 
                                  &                                                                                    &                                                                                  & Test               & Val              & M-avg         & Short    & Medium    & Long    & All    & Short     & Medium     & Long     & All     \\ \midrule
\rowcolor{gray!20} LLaVA-Video 7B & 64frm                                                                              & 13k                                                                              & 65.0               & 42.6
& 71.2          & 76.1     & 61.0      & 52.4    & 63.2   & 71.5      & 60.7       & 52.1     & 60.0    \\ \midrule
AKS~\cite{TangQXTJY25}            & 1fps                                                                               & 13k                                                                              & 85.4                   & 47.4                 & 72.0              & \textbf{77.2}& \textbf{64.8}& 53.9    & \textbf{65.3}& \textbf{72.3}& \underline{62.1}& \textbf{57.4}& \underline{62.7}\\
Panels~\cite{abs-2509-23724}      & 1fps                                                                               & 13k                                                                              & 79.2               & -                & -             & -        & 62.2      & \underline{54.0}& 64.4   & -         & -          & -        & -       \\
DToMA~\cite{YuanYB25}             & -                                                                                  & 12k                                                                              & -                  & -                & 71.7          & -        & -         & -       & \underline{65.0}& -         & -          & -        & 59.6    \\
Video-RAG~\cite{abs-2411-13093}   & -                                                                                  & 15k                                                                              & -                  & -                & \textbf{72.4}& -        & -         & -       & -      & -         & -          & -        & 58.7    \\
AdaRETAKE~\cite{WangSZWCN25}      & 1024frm                                                                            & 40k                                                                              &                    \textbf{86.2}& \underline{49.6}& 71.7          &          75.8
&           62& 52.9& 63.6&           69.7&            59.2&          52.8& 59.4\\
\textbf{$\text{\ours}^*$}           & 512/1024frm                                                                        & 13k                                                                              & \underline{85.9}& \textbf{51.0}& \textbf{72.4}& \underline{76.3}& \underline{63.3}& \textbf{54.4}& 64.7   & \underline{71.5}& \textbf{65.5}& \underline{57.3}& \textbf{63.6}\\ \midrule
\rowcolor{gray!20} LLaVA-OV 7B    & 32frm                                                                              & 7k                                                                               & 56.3               & 38.4& 63.4          & \underline{70.6}& 54.8      & 48.2    & 57.8   & \textbf{69.5}& 53.4       & 49.8     & 56.2    \\ \midrule
AKS~\cite{TangQXTJY25}            & 1fps                                                                               & 7k                                                                               &                    &                  \underline{43.5}&               \underline{68.3}&          &           &         & 58.4   &           65.9&            \textbf{58.9}&          \underline{54.3}& \underline{58.9}\\
Panels~\cite{abs-2509-23724}      & 1fps                                                                               & 7k                                                                               & 69.5               & -                & -             & -        & 56.2      & \underline{50.2}& 58.9   & -         & -          & -        & -       \\
BOLT~\cite{00010XG25}             & 1fps                                                                               & 7k                                                                               & -                  & -                & 65.8          & 69.2     & \underline{56.8}& 47.3    & 57.8   & -         & -          & -        & 57.0    \\
AdaRETAKE~\cite{WangSZWCN25}      & 1024frm                                                                            & 20k                                                                              &                    \underline{75.8}& 42.1             & 64.4          & \textbf{72.1}& 53.6      & \textbf{51.4}& \textbf{59.0}& \underline{68.5}& 51.0       & 47.2     & 54.2    \\
\textbf{$\text{\ours}^*$}           & 512/1024frm                                                                        & 7k                                                                               & \textbf{80.5}& \textbf{46.2}& \textbf{68.9}& 70.0     & \textbf{57.3}& 49.8    & \textbf{59.0}& 67.9      & \underline{58.0}& \textbf{55.0}& \textbf{59.4}\\ \bottomrule
\end{tabular}
}
\vspace{-5pt}
\end{table*}

\begin{table*}[t]
\centering
\caption{Performance comparison of \ours against representative video RAG method (AKS) and visual compression methods based on LLaVA-Video across five long VideoQA benchmarks. All methods runs on a single 3090 with the fully use of memory overhead.}
\vspace{-5pt}
\label{tb:overall_limited}
\resizebox{\linewidth}{!}{
\renewcommand{\arraystretch}{0.8}
\begin{tabular}{c|c|c|c|c|c|cccc|cccc}
\toprule
\multirow{2}{*}{\textbf{Method}} & \multirow{2}{*}{\textbf{\begin{tabular}[c]{@{}c@{}}Sampled\\ Frames\end{tabular}}} & \multirow{2}{*}{\textbf{\begin{tabular}[c]{@{}c@{}}Input\\ Tokens\end{tabular}}} & \textbf{TimeScope} & \textbf{LVBench} & \textbf{MLVU} & \multicolumn{4}{c|}{\textbf{Video-MME}}                       & \multicolumn{4}{c}{\textbf{LongVideoBench}}                   \\ \cmidrule(l){4-14} 
                                 &                                                                                    &                                                                                  & Test               & Val              & M-avg         & Short         & Medium        & Long          & All           & Short         & Medium        & Long          & All           \\ \midrule
LLaVA-Video 7B                   & 32frm                                                                              & 7k                                                                               & 58.3               & 41.4             & 68.5          & \underline{74.8}    & 58.4          & 52.0          & 61.7          & \underline{70.6}    & 59.0          & 50.5          & 58.6          \\ \midrule
AKS                              & 1fps                                                                               & 7k                                                                               & \underline{84.6}         & 46.6             & 70.8          & 74.7          & \underline{61.9}    & 51.7          & 62.8          & 68.1          & 60.0          & \underline{54.1}    & 59.7          \\
AdaRETAKE                        & 384frm                                                                             & 40k                                                                              & 78.2               & \underline{46.8}       & \underline{71.7}    & 74.7          & 61.8          & \underline{54.3}    & \underline{63.6}    & 69.0          & \underline{60.4}    & 53.5          & \underline{59.8}    \\
\textbf{$\text{\ours}$}          & 512/1024frm                                                                        & 13k                                                                              & \textbf{85.6}      & \textbf{50.2}    & \textbf{72.3} & \textbf{76.2} & \textbf{62.7} & \textbf{55.0} & \textbf{64.6} & \textbf{71.5} & \textbf{63.6} & \textbf{57.2} & \textbf{63.0} \\ \bottomrule
\end{tabular}
}
\vspace{-10pt}
\end{table*}

Since the attention scores received by visual tokens are generally uniform in shallow layers~\cite{XingHDL0ZCHWWL25, abs-2501-12386}, we only leverage the attention weights from deeper layers to calculate relevance scores in practice, \emph{e.g.}, after the $2$-th layer.

\subsubsection{Fast Memory Indexing}
\label{sec:memidx}

Although the encoding-based reading solution can accurately capture the video-question similarity based on MLLMs, its practical use is still limited due to the repeated MLLM inference for new questions. In this case, we also explore the fast memory index method, termed \emph{MemIndex}.

In terms of fast and flexible memory retrieval, we assume that MemIndex should has the following properties. 
First, MemIndex should be independent to the encoding of visual memory, thus they can efficiently handle multiple questions or streaming cases~\cite{NiuLMGZHDDD0ZZC25, abs-2411-03628}. 
Second, the index features of MemIndex should be compact enough, either for the visual or the question ones, thereby further reducing the cost of cross-modal matching.

Achieving the above target is still intractable. For instance, the offline memory caches and the question ones still have a certain semantic gap~\cite{abs-2508-07675}, although they are encoded by the same MLLMs. Besides, the computation of retrieval is still expensive, even using the compressed cache tokens, \emph{i.e.}, 21$k$ cache tokens of 25 layers.

To this end, we first consider the encoding-based reading as the upper-bound of MemIndex, and then the objective of MemIndex is defined by
\begin{equation}
\label{eq:fast_obj}
\resizebox{0.85\columnwidth}{!}{$\displaystyle{
\arg\min_{\sigma} \sum_{i=1}^D \left\| \sigma(R_i)  - g_i \right\|_2, \: \text{where} \: \sigma(R_i)=\sum_{l=3}^L \alpha^l r_i^l.
}$}
\end{equation}
Here, $D$ is the number of training data used for optimization, and $r_i^l$ is the relevance score of clip $V_i$ in the $l$-th layer obtained from MemIndex. We aim to find a linear regression function $\sigma(\cdot)$ that minimizes the L2 distance to the ``target'' score $g_i$ in Eq.~\ref{eq:s_ret}.

Specifically, given the input question $T_q$, we first encode its features via the MLLM, denoted as $\mathbf{Q}_{T_q}$. Then, the basic VL matching can be defined by 
\begin{equation}
    \centering
    \begin{aligned}
        & \mathbf{A}^l_c = \text{Attention}(\mathbf{Q}_{T_q}, \mathbf{\hat{K}}_{V_i}), \\
        & r_i = \sum_{l=3}^L \sum_{j \in T_q}  \sum_{k \in V_i} r_{jk}^l.
    \end{aligned}
\end{equation}
where $\mathbf{\hat{K}}_{V_i}$ is the compressed key vectors of clip $V_i$ in the stored memory $M_i$. 

Although feasible, this basic solution still involves excessive visual and text tokens of all layers. In this case, we first conduct the selection of visual cache layers according to the fitted regression function $\sigma(\cdot)$ in Eq.~\ref{eq:fast_obj}:
\begin{equation}
    \mathcal{H} = \{l | \alpha^l \in \text{top-}K(\{\alpha^l\}_{l=3}^L)\},
\end{equation}
where $K$ is the number of selected representative cache layers. 
We identify these cache layers with highest learned weights $\alpha^l$, which naturally indicate each layer's importance for relevance computation.

Besides, we also revise FlexMem during the memory encoding, using a higher-ratio of compression to obtain more compact local memories as the visual index tensor, \emph{e.g.}, the size can be changed from $|I_i|\times \frac{M}{N} \times d$ to $k \times d$. 
In terms of the question tokens, we empirically select the last token as the index feature~\cite{abs-2503-02656,abs-2410-17247}. In this case, the index of FlexMem can be defined by 
\begin{equation}
    \begin{aligned}
        \mathbf{q} &= \mathbf{Q}_{T_q}[-1],\: \mathbf{K}^*_{V_i}=\{\mathbf{k}_j^l|\hat{s}_j^l \in \overset{k}{\underset{j\in V_i}{\arg\max}} \: \hat{s}_j^l\}, \\
        \mathbf{\hat{A}}^l_c &= \text{Attention}(\mathbf{q}, \mathbf{K}^*_{V_i}), \\
        \hat{r}_i &= \sum_{l\in \mathcal{H}}  \sum_{j \in V_i^*} \hat{r}_{j}^l.
    \end{aligned}
\end{equation}
Here $k$ is the number of key vectors selected as the representative visual indexes.

\begin{table*}[t]
\centering
\caption{Comparison between SOTA Video-MLLMs and LLaVA-Video with FlexMem on five long VideoQA benchmarks.}
\vspace{-5pt}
\label{tb:overall_sota}
\resizebox{\linewidth}{!}{
\renewcommand{\arraystretch}{0.8}
\begin{tabular}{c|c|c|c|c|cccc|ccc}
\toprule
\multirow{2}{*}{\textbf{Method}}   & \multirow{2}{*}{\textbf{LLM}} & \textbf{TimeScope}         & \textbf{LVBench}           & \textbf{MLVU}              & \multicolumn{4}{c|}{\textbf{Video-MME}}                                                                           & \multicolumn{3}{c}{\textbf{LongVideoBench}}                                          \\ \cmidrule(l){3-12} 
                                   &                               & Test                       & Val                        & M-avg                      & Short                      & Medium                     & Long                       & All                        & Medium                     & Long                       & All                        \\ \midrule
\rowcolor{gray!20} GPT-5           & -                             & -                          & -                          & 77.3                       & -                          & -                          & -                          & 81.8                       & -                          & -                          & 72.6                       \\
\rowcolor{gray!20} GPT-4o          & -                             & -                          & 27.0                       & 64.6                       & 80.0                       & 70.3                       & 65.3                       & 71.9                       & 69.1                       & 60.9                       & 66.7                       \\
\rowcolor{gray!20} Gemini-1.5-Pro  & -                             & -                          & 33.1                       & -                          & 81.7                       & 74.3                       & 67.4                       & 75.0                       & 65.3                       & 58.6                       & 64.0                       \\ \midrule
Video-XL~\cite{ShuLZQ0L0025}       & 7B                            & -                          & -                          & 64.9                       & 62.0                       & 53.2                       & 49.2                       & 55.5                       & 49& 45.2& 50.5\\
mPLUG-Owl3~\cite{Ye0LH0000025}     & 7B                            & -                          & 43.5                       & 63.7                       & 70.0                       & 57.7                       & 50.1                       & 59.3                       & -                          & -                          & 52.1                       \\
Qwen2.5-VL~\cite{Qwen2.5-VL}       & 7B                            & \underline{81.0}                 & \underline{45.3}                 & 70.2                       & -                          & -                          & -                          & \underline{65.1}                 & -                          & -                          & 56.0                       \\
TimeMarker~\cite{abs-2411-18211}   & 8B                            & -                          & 41.3                       & 63.9                       & 71.0                       & 54.4                       & 46.4                       & 57.3                       & -                          & -                          & 56.3                       \\
LongVU~\cite{abs-2410-17434}       & 7B                            & -                          & -                          & 65.4                       & -                          & -                          & \textbf{59.5}              & 60.6                       & -                          & -                          & -                          \\
TSPO~\cite{abs-2508-04369}         & 7B                            & -                          & \underline{45.3}                 & \textbf{76.3}              & -                          & -                          & 54.7                       & \textbf{65.5}              & -                          & -                          & \textbf{63.9}              \\
LongVA~\cite{ZhangZLZYZWTLL25}     & 7B                            & 55.9                       & -                          & 56.3                       & 61.1                       & 50.4                       & 46.2                       & 52.6                       & -                          & -                          & -                          \\
ByteVideoLLM~\cite{abs-2412-09530} & 14B                           & -                          & -                          & 70.1                       & 74.4                       & \underline{62.9}                 & \underline{56.4}                 & 64.6                       & -                          & -                          & -                          \\ \midrule
\textcolor{black!50}{LLaVA-Video}  & \textcolor{black!50}{7B}      & \textcolor{black!50}{65.0} & \textcolor{black!50}{42.1} & \textcolor{black!50}{71.2} & \textcolor{black!50}{\underline{76.1}} & \textcolor{black!50}{61.0} & \textcolor{black!50}{52.4} & \textcolor{black!50}{63.2} & \textcolor{black!50}{\underline{60.7}} & \textcolor{black!50}{\underline{52.1}} & \textcolor{black!50}{60.0} \\
\textbf{+ $\text{\ours}$}          & 7B                            & \textbf{85.9}              & \textbf{51.0}              & \underline{72.4}                 & \textbf{76.3}              & \textbf{63.3}              & 54.4                       & 64.7                       & \textbf{65.5}              & \textbf{57.3}              & \underline{63.6}                 \\ \bottomrule
\end{tabular}

}
\vspace{-10pt}
\end{table*}

\begin{table}[h]
\centering
\caption{
Comparison of our method, LLaVA-Video integrated with FlexMem and MemIndex, with SOTA online and offline models on backward tracing task of OVOBench.
EPM, ASI and HLD denote \emph{EPisodic Memory}, \emph{Action Sequence Identification} and \emph{HaLlucination Detection}, respectively.
}
\vspace{-5pt}
\label{tb:overall_stream}
\resizebox{\columnwidth}{!}{
\renewcommand{\arraystretch}{0.8}
\begin{tabular}{c|c|c|cccc}
\toprule
\multirow{2}{*}{\textbf{Methods}}      & \multirow{2}{*}{\textbf{LLM}} & \multirow{2}{*}{\textbf{\# Frames}} & \multicolumn{4}{c}{\textbf{Backward Tracing}}                                                                     \\ \cmidrule(l){4-7} 
                                       &                               &                                     & EPM                        & ASI                        & HLD                        & Average                    \\ \midrule
\multicolumn{7}{c}{\textbf{Offline Models}}                                                                                                                                                                                      \\ \midrule
\textcolor{black!50}{Gemini-1.5-Pro}   & \textcolor{black!50}{-}       & \textcolor{black!50}{-}             & \textcolor{black!50}{58.6} & \textcolor{black!50}{76.4} & \textcolor{black!50}{52.6} & \textcolor{black!50}{62.5} \\
InternVL-V2~\cite{abs-2412-05271}                    & 8B                            & 64                                  & \underline{48.2}                 & 57.4                       & \underline{24.7}                 & \underline{43.4}                 \\
LongVU~\cite{abs-2410-17434}           & 7B                            & 1fps                                & 40.7                       & \underline{59.5}                 & 4.8                        & 35.0                       \\ \midrule
\multicolumn{7}{c}{\textbf{Online Models}}                                                                                                                                                                                       \\ \midrule
Flash-VStream~\cite{abs-2406-08085}    & 7B                            & 1fps                                & 39.1                       & 37.2                       & 5.9                        & 27.4                       \\
VideoLLM-online~\cite{ChenLWLSGLGMS24} & 8B                            & 2fps                                & 22.2                       & 18.8                       & \underline{12.2}                 & 17.7                       \\
Dispider~\cite{0001DD0ZCL025}          & 7B                            & 1fps                                & \underline{48.5}                 & \underline{55.4}                 & 4.3                        & \underline{36.1}                 \\ \midrule
LLaVA-Video                            & 7B                            & 64                                  & 55.2                       & \textbf{60.8}              & 42.5                       & 52.8                       \\
\textbf{+\ours w. MemIndex}            & 7B                            & 1fps                                & \textbf{57.6}              & 54.1              & \textbf{49.5}              & \textbf{54.4}              \\ \bottomrule
\end{tabular}

}
\vspace{-10pt}
\end{table}

\section{Experiment}
\label{sec:exp}

\subsection{Benchmarks and Metrics}

To validate \ours, we conduct extensive experiments on five benchmarks for long video understanding, including MLVU~\cite{abs-2406-04264}, LongVideoBench~\cite{WuLCL24}, LVBench~\cite{abs-2406-08035}, Video-MME~\cite{FuDLLRZWZSZCLLZ25} and TimeScope~\cite{ZoharWD0XHYWJZY25}. 
MLVU includes videos ranging from 3 minutes to 2 hours that require comprehensive temporal understanding.
Video-MME covers videos of diverse genres and durations, including short, medium, and long-form content.
LongVideoBench is designed for tasks requiring precise retrieval and reasoning over detailed multimodal information within extended temporal contexts, containing videos up to an hour in length.
LVBench challenges MLLMs to demonstrate long-term memory retention and extended comprehension capabilities, with an average video duration of approximately 68.4 minutes.
TimeScope probes the limits of long video capabilities with videos ranging from 1 minute to 8 hours.

\subsection{Implementation Details}
 
\ours is designed as a training-free approach that can be seamlessly integrated with existing MLLMs without requiring additional fine-tuning. We validate \ours using two recent MLLMs: LLaVA-Video~\cite{ZhangWLLMLL25} and LLaVA-OneVision~\cite{0080ZGZ00ZZL0L25}.
we evaluate the effectiveness of \ours on long VideoQA tasks through \emph{encoding-based reading}, and equip \ours with \emph{MemIndex} in streaming QA tasks.
We uniformly sample $512$ frames on TimeScope, LVBench, and MLVU, while sampling $1024$ frames on Video-MME and LongVideoBench. 
The input token counts for final decoding are $13k$ and $7k$ for LLaVA-Video and LLaVA-OneVision respectively, maintaining consistency with their corresponding baselines using sparse uniform sampling strategies. For our MemIndex implementation, we select $K=3$ visual cache layers and $k=5$ visual index features to enable efficient memory indexing while preserving representative information.

\subsection{Quantitative Analysis}

\noindent\textbf{Comparison with existing methods.} 
Tab.~\ref{tb:overall_method} presents a comprehensive comparison of \ours against representative VideoRAG and visual compression methods across two recent MLLMs, \emph{i.e.}, LLaVA-Video~\cite{ZhangWLLMLL25} and LLaVA-OneVision~\cite{0080ZGZ00ZZL0L25}.
From Tab.~\ref{tb:overall_method}, we can first observe that existing methods typically require dense frame sampling and numerous token inputs for final decoding. 
VideoRAG methods like AKS excel at visual evidence localization on LongVideoBench, and visual compression methods like AdaRETAKE demonstrate strong holistic video understanding on Video-MME.
In contrast, \ours consistently enhances the performance of both base models, achieving SOTA results against other methods built upon the same MLLMs across most benchmarks. 
This demonstrates \ours's effectiveness in comprehensive memory construction through iterative processing and precise backward tracing via memory recall.
For instance, \ours enables LLaVA-Video to surpass its baseline by 32.2\% on TimeScope and 19.7\% on LVBench.
These results conclusively validate the effects of \ours in advancing long video comprehension capabilities of MLLMs.

\noindent\textbf{Comparison with limited memory overhead.} 
We evaluate the scalability and performance gains of \ours compared to two representative methods on \textbf{a single 3090 GPU}, \emph{i.e.}, AdaRETAKE~\cite{WangSZWCN25}, which exemplifies visual compression approaches, and AKS~\cite{TangQXTJY25}, representing VideoRAG methods. 
As shown in Tab.~\ref{tb:overall_limited}, we first observe that AdaRETAKE and AKS experience considerable degradation compared to their unrestricted performance in Table~\ref{tb:overall_method}.
For instance, when GPU memory budget is limited to 24GB, the input capacity of AdaRETAKE is reduced from $1024$ to $384$ frames, and its performance drops by an average of 3.3\% across all benchmarks.
In contrast, \ours consistently maintains superior performance under resource constraints, and retains 99.5\% of its full performance.
Overall, these results demonstrating \ours's ability to flexibly manage visual memories while preserving essential information.

\noindent\textbf{Comparison with SOTA Video-MLLMs.} 
We further compare \ours with existing SOTA Video-MLLMs on five benchmarks in Tab.~\ref{tb:overall_sota}. As shown in Tab.~\ref{tb:overall_sota}, when employing the uniform sampling strategy, short Video-MLLMs such as Qwen2.5-VL exhibit superior performance on Video-MME requiring global understanding capabilities. However, this straightforward solution significantly underperforms compared to visual compression methods like TSPO on LongVideoBench, which requires fine-grained detail reasoning over extended video durations.
We can also see that \ours consistently achieves competitive or superior performance compared to other methods with comparable model sizes. Notably, \ours improves LLaVA-Video to the level of Gemini-1.5-Pro, while significantly surpassing it by 54.1\% on LVBench. 
Overall, these results well confirm the effectiveness of our \ours in improving long video understanding of MLLMs.

\begin{table}[t]
\centering
\caption{Ablation studies on different designs of FlexMem under the encoding-based reading setting across two benchmarks. Methods marked with $\ddag$ indicate our chosen settings.}
\vspace{-5pt}
\label{tb:ab_slow}
\resizebox{\columnwidth}{!}{
\renewcommand{\arraystretch}{0.8}
\begin{tabular}{c|cccc|c}
\toprule
                                   & \multicolumn{4}{c|}{\textbf{LongVideoBench}}                                                                           & \textbf{LVBench} \\ \cmidrule(l){2-6} 
\multirow{-2}{*}{\textbf{Choices}} & Short                       & Medium                      & Long                        & All                         & Val              \\ \midrule
\multicolumn{6}{c}{\textbf{Compression Strategy}}                                                                                                                             \\ \midrule
Context Compression Only         & 70.1                        & 64.3                        & 56.2                        & 62.5                        & 49.9             \\
Local Compression Only              & 70.9                        & 64.6                        & 55.9                        & 62.6                        & 49.7             \\
Dual-Pathway$^{\ddag}$             & 71.5                        & 65.5                        & 57.3                        & 63.6                        & 51.0             \\ \midrule
\multicolumn{6}{c}{\textbf{Context during Prefill}}                                                                                                                            \\ \midrule
Context Memory Only                & 71.2                        & 65.0                        & 53.5                        & 61.9                        & 50.5             \\
Local Memory Only                  & 71.2                        & 63.3                        & 54.6                        & 61.8                        & 50.0             \\
Combination of both$^{\ddag}$ & 71.5                        & 65.5                        & 57.3                        & 63.6                        & 51.0             \\ \midrule
\multicolumn{6}{c}{\textbf{Context during Decoding}}                                                                                                                          \\ \midrule
All $M_{bank}$               & {\color[HTML]{333333} 71.5} & {\color[HTML]{333333} 58.7} & {\color[HTML]{333333} 53.2} & {\color[HTML]{333333} 59.8} & 49.3                 \\
Memory Reading$^{\ddag}$   & 71.5                        & 65.5                        & 57.3                        & 63.6                        & 51.0             \\ \midrule
\multicolumn{6}{c}{\textbf{Number of Frames in Each Clip}}                                                                                                                                       \\ \midrule
8$^{\ddag}$                        & 71.5                        & 65.5                        & 57.3                        & 63.6                        & 51.0             \\
16                                 & 71.5                        & 64.8                        & 57.1                        & 63.4                        & 50.1             \\
32                                 & 70.1                            & 62.6                            & 55.7                            & 61.7                            & 49.3                 \\ \bottomrule
\end{tabular}
}
\vspace{-5pt}
\end{table}

\begin{table}[t]
\centering
\caption{Ablation studies on index token designs of \ours with MemIndex on two benchmarks. \emph{Single} and \emph{Multi} denote the Single-Detail and Multi-Detail tasks on MLVU, respectively. \emph{AttEnc} means token selection based on local saliency score.}
\vspace{-5pt}
\label{tb:ab_fast}
\resizebox{\columnwidth}{!}{
\begin{tabular}{ccc|cccc|c}
\toprule
\multirow{2}{*}{\textbf{Layers}} & \multirow{2}{*}{\textbf{Text}} & \multirow{2}{*}{\textbf{Vision}} & \multicolumn{4}{c|}{\textbf{MLVU}} & \textbf{LVBench}     \\ \cmidrule(l){4-8} 
                                 &                                &                                  & Single & Multi & Holistic & M-avg & Val                  \\ \midrule
\multicolumn{3}{c|}{Encoding-based Index}                                                                          & 77.1   & 54.8  & 77.3     & 72.4  & 51.0                 \\ \midrule
All                               & All                            & All                              & 76.9   & 54.0  & 77.1     & 72.0  & 46.3                     \\
3                                & All                            & All                              & 77.2   & 53.3  & 77.5     & 72.2  & 46.6 \\
3                                & Last-Token                     & All                              & 77.4   & 53.3  & 77.3     & 72.3  & 46.8                     \\
3                                & Last-Token                     & AttEnc                       & 77.1   & 53.1  & 77.5     & 72.1  & 45.7                     \\ \bottomrule
\end{tabular}

}
\vspace{-10pt}
\end{table}

\begin{figure*}[ht]
  \centering
   \includegraphics[width=\linewidth]{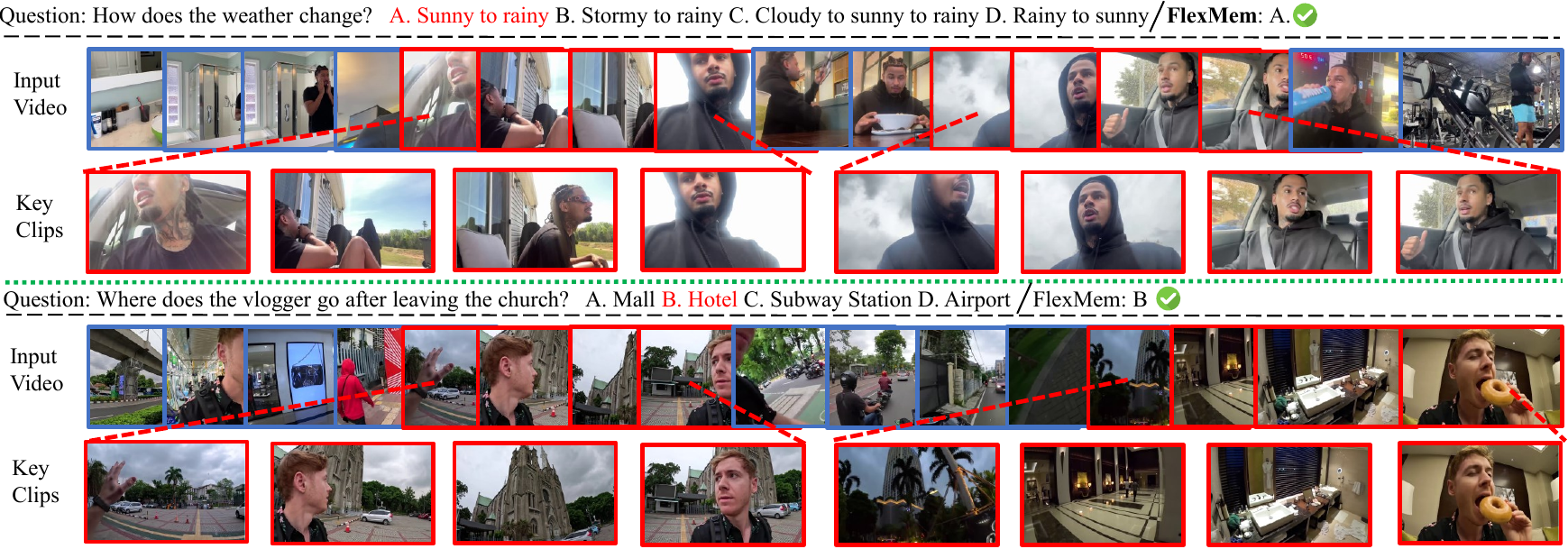}
   \caption{Qualitative evaluation of \ours. Input Video denote the sampled frames, and Key Fragments are the selected clips for answer generation via memory recall mechanism. These results demonstrate \ours's capacity in comprehensive and fine-grained visual understanding.}
   \label{fig:quali_ana}
    \vspace{-10pt}
\end{figure*}

\noindent\textbf{Results of \ours + MemIndex on streaming QA task.} 
Tab.~\ref{tb:overall_stream} compares the performance of \ours integrated with MemIndex against existing SOTA online and offline models in streaming QA tasks.
As shown in Tab.~\ref{tb:overall_stream}, we can see that while offline models such as LongVU exhibit superior holistic comprehension capabilities compared to online methods like Dispider on ASI, their performance degraded on EPM that requires historical memory localization. 
After equipped with FlexMem and MemIndex, LLaVA-Video exceeds its common version by 3\% on average, demonstrating the capacity of our method to effective memory recall and flexible context management.
Overall, these results show the merits of MemIndex in historical information tracing.

\noindent\textbf{Ablation Study.} Here, we first ablate the key designs choices of \ours in Tab.~\ref{tb:ab_slow}. 
In the first block of Tab.~\ref{tb:ab_slow}, we examine the effects of our dual-pathway compression strategy. \emph{Context Compression Only} and \emph{Local Compression Only} denote memory compression using only $s_j^l$ in Eq.~\ref{eq:s_pre} and $\hat{s}_j^l$ in Eq.~\ref{eq:s_dec}, respectively. 
The results show that the context features can transfer historical information for long video understanding, while local features effectively compress memories on short videos. 
Notably, the performance gains of our \emph{Dual-Pathway} become more pronounced with longer video durations, validating its ability to effectively exploit the distinct roles of MLLMs during prefill and decoding phases, \emph{i.e.}, encoding clips with contextual memories and generating predictions with stored memories.

In the second block of Tab.~\ref{tb:ab_slow}, we validate the effectiveness of context memory and local memory during the prefill stage.
We observe that while employing either context memory or local memory alone during clip encoding yields reasonable performance, their combination results in significantly enhanced performance. This finding indicates that the two memory types are complementary, \emph{i.e.}, context memory maintains temporal continuity while local memory preserves long-range dependencies.
The third block examines the benefits of our memory reading strategy compared to indiscriminate loading of all memory.
The results demonstrate that our memory recall can effectively identify and prioritize a small subset of key clips from extended videos.
In the last block of Tab.~\ref{tb:ab_slow}, we analyze performance across different block sizes. The results indicate that MLLMs consistently require detailed visual information through smaller block sizes, regardless of video duration.
Overall, these results further confirm the effectiveness of our proposed designs choices for \ours.

Next, we further ablate the effectiveness of our fast memory indexing discussed in Sec.~\ref{sec:memidx}, as shown in Tab.~\ref{tb:ab_fast}.
The most simple solution is computing relevance scores across all cache layers for all visual KVs, which inevitably introduces substantial computational overhead and information redundancy. 
In contrast, our MemIndex achieves comparable or even superior performance on MLVU compared to encoding-based index while significantly reducing computational complexity.
Overall, the results demonstrate that our MemIndex substantially reduces computational costs with minimal performance degradation.

\subsection{Qualitative Analysis}

In Fig.~\ref{fig:quali_ana}, we visualize the comprehensive long video understanding and precise memory recall capabilities of \ours. As observed, \ours can significantly improve the baseline MLLM for long video understanding through precise visual evident localization. 
While sparse uniform sampling strategies typically lead to poor performance in long video comprehension, \ours empowers MLLMs to iteratively process entire videos and generate accurate answers via precise memory recall.

\section{Conclusion}

In this paper, we presented FlexMem, a novel training-free approach that enables MLLMs to understand videos of infinite lengths via a flexible visual memory mechanism. FlexMem iteratively processes video content and recalls key memory fragments for question answering, breaking the input length limitations of MLLMs.
Notably, FlexMem achieves substantial performance gains over two representative methods on a single 3090, and enables MLLMs to achieve comparable or superior performance to SOTA models like GPT-4o on several benchmarks.

\section{Acknowledgments}

This work is supported by the National Key Research and Development Program of China (No. 2025YFE0113500), the National Science Fund for Distinguished Young Scholars (No. 62525605), the National Natural Science Foundation of China (No. U25B2066, No. U22B2051, No.62572407) , Fujian Province Special Science and Technology Program (No. 2025H0041).

{
    \small
    \bibliographystyle{ieeenat_fullname}
    \bibliography{main}
}

\end{document}